\documentclass[10pt,twocolumn,letterpaper]{article}

\usepackage{cvpr}
\usepackage{times}
\usepackage{epsfig}
\usepackage{graphicx}
\usepackage{amsmath}
\usepackage{amssymb}
\usepackage{subcaption}

\usepackage{diagbox}

\usepackage[pagebackref=true,breaklinks=true,letterpaper=true,colorlinks,bookmarks=false]{hyperref}

\usepackage{xcolor}

\definecolor{purple}{rgb}{1,0,1}
\newcommand{\norm}[1]{\left\lVert#1\right\rVert}

\cvprfinalcopy 

\def\cvprPaperID{4536} 

\ifcvprfinal\fi
\begin{document}

\title{Unsupervised Multi-Modal Image Registration via Geometry Preserving Image-to-Image Translation}

\author{
Moab Arar\\
Tel Aviv University
\and 
Yiftach Ginger\\
Tel Aviv University 
\and 
Dov Danon\\
Tel Aviv University
\and 
Ilya Leizerson\\
Elbit Systems
\and 
Amit H. Bermano\\
Tel Aviv University
\and
Daniel Cohen-Or\\
Tel Aviv University}

\maketitle

\begin{abstract}
Many applications, such as autonomous driving, heavily rely on multi-modal data where spatial alignment between the modalities is required. Most multi-modal registration methods struggle computing the spatial correspondence between the images using prevalent cross-modality similarity measures. In this work, we bypass the difficulties of developing cross-modality similarity measures, by training an image-to-image translation network on the two input modalities. This learned translation allows training the registration network using simple and reliable mono-modality metrics. We perform multi-modal registration using two networks - a spatial transformation network and a translation network. We show that by encouraging our translation network to be geometry preserving, we manage to train an accurate spatial transformation network. Compared to state-of-the-art multi-modal methods our presented method is unsupervised, requiring no pairs of aligned modalities for training, and can be adapted to any pair of modalities. We evaluate our method quantitatively and qualitatively on commercial datasets, showing that it performs well on several modalities and achieves accurate alignment.
\end{abstract}


\section {Introduction}
\begin{figure}
    \centering
    \includegraphics[width=0.46\textwidth]{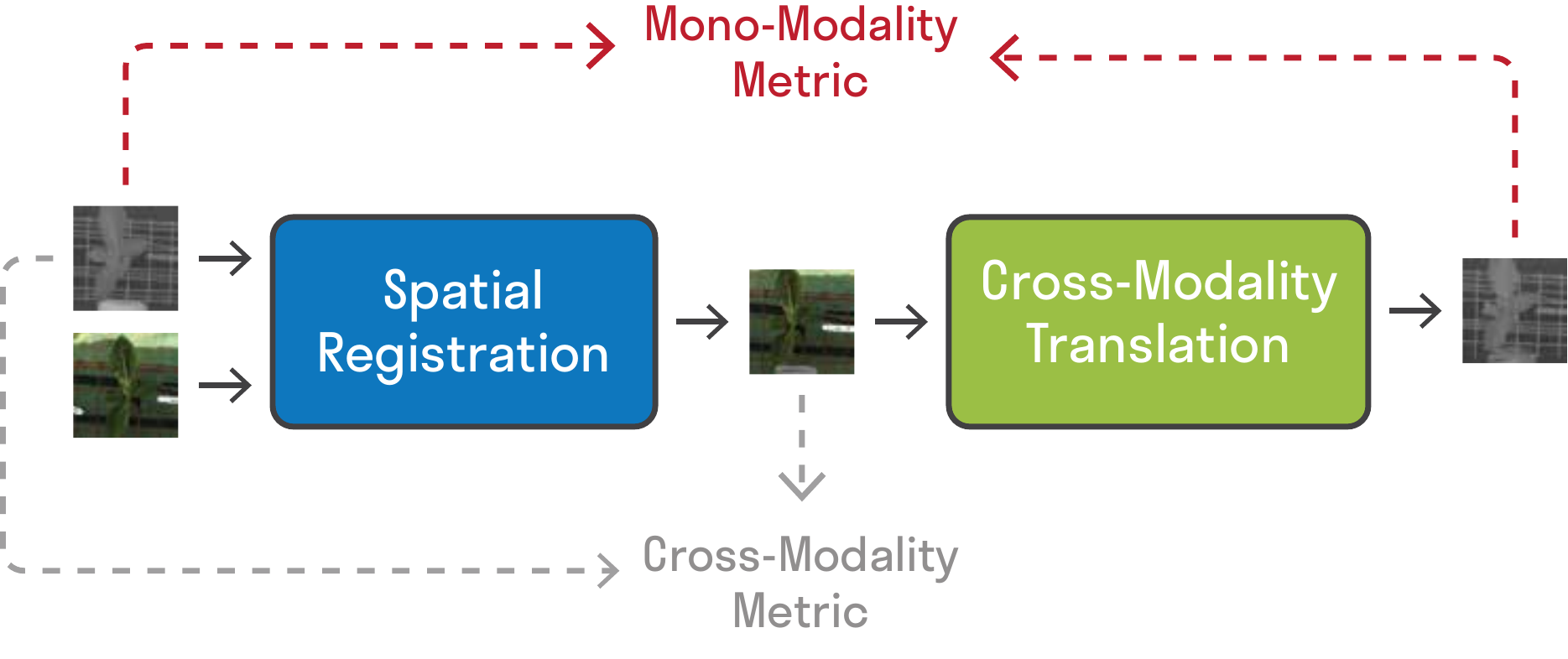}
    \caption{\textbf{Method overview.} Conventional methods (faded dashed at the bottom) use cross-modality metrics (e.g., Normalized Cross Correlation) to optimize a spatial transformation function. Our method learns a cross-modality translation, mapping between the two modalities. This enables the use of a reliable accurate mono-modality metric instead.}
    
\label{fig:NeMAR}
\end{figure}

 Scene acquisition using different sensors is common practice in various disciplines, from classical ones such as medical imaging and remote sensing, to emerging tasks such as autonomous driving. Multi-modal sensors allow gathering a wide range of physical properties, which in turn yields richer scene representations. For example, in radiation planning, multi-modal data (e.g. Computed Tomography (CT) and Magnetic Resonance Imaging (MRI) scans) is used for more accurate tumor contouring which reduces the risk of damaging healthy tissues in radiotherapy treatment~\cite{oh2017deformable,schmidt2015radiotherapy}. More often than not, multi-modal sensors naturally have different extrinsic parameters between modalities, such as lens parameters and relative position. In these cases, non-rigid image registration is essential for proper execution of the aforementioned downstream tasks. 
 
 Classic \textit{multi-modal image registration} techniques attempt to warp a source image to match a target one via a non-linear optimization process, seeking to maximize a predefined similarity measure~\cite{ZITOVA2003977}. Besides a computational disadvantage, which is critical for applications such as autonomous driving, effectively designing similarity measures for such optimization has proven to be quite challenging. This is true for both intensity-based measures, commonly used in the medical imaging~\cite{haskins2019deep}, and feature-based ones, typically adapted for more detailed modalities (e.g Near Infra-Red (NIR) and RGB)~\cite{ShenXZJ14}.
 

 These difficulties gave rise to the recent development of deep regression models. These types of models typically have lengthy training time, either supervised or unsupervised, yet they offer expeditious inference that usually generalizes well. Since it is extremely hard to collect ground-truth data for the registration parameters, supervised multi-modal registration methods commonly use synthesized data in order to train a registration network~\cite{RegNet, Zampieri2018MultimodalIA}. This makes their robustness highly dependent on the similarity between the artificial and real-life data distribution and appearance. Unsupervised registration techniques, on the other-hand, frequently incorporate a spatial transform network (STN)~\cite{STN} and train an end-to-end network \cite{unsupervised_stn_mono_2_jointloss, unsupervised_stn_mono, unsupervised_stn_gan_mono_1_jointloss, unsupervised_stn_gan_multi_1_jointloss, unsupervised_stn_gan_multi_3_justgan}. 
 
 Typically, such approaches optimize an STN by comparing the deformed image and the target one using simple similarity metrics such as pixel-wise Mean Squared Error (MSE) \cite{unsupervised_fcn_deformation_1, MSE1, MSE2}. Of course, such approaches can only be used in mono-modality settings and become irrelevant for multi-modality settings. To overcome this limitation, unsupervised multi-modal registration networks use statistics-based similarity metrics, particularly, (Normalized) Mutual Information ((N)MI) \cite{NMI_SSIM}, Normalized Cross Correlation (NCC) \cite{NCC1}, or Structural Similarity Index Metric (SSIM) \cite{elastic_registration_nostn_multi_justgan, NMI_SSIM} (see Figure~\ref{fig:NeMAR}, faded dashed path). However, these metrics are either computationally intractable (e.g., MI)~\cite{pmlr-v80-belghazi18a} and hence cannot be used in gradient-based methods, or are domain-dependent (e.g., NCC), failing to generalize for all modalities.
 
In this paper, we present an unsupervised method for multi-modal registration. In our work, we exploit the celebrated success of Multi-Modal Image Translation~\cite{pix2pix2016, CycleGAN2017, zhu2017toward, huang2018munit}, and simultaneously learn multi-modal translation and spatial registration. The key idea is to alleviate the shortcomings of a hand-crafted similarity measure by training an image-to-image translation network $T$ on two given modalities. This in turn will let us use mono-modality metrics for evaluating our registration network $R$ (see Figure~\ref{fig:NeMAR}, vivid path on the top). 

The main challenge for this approach is to train the registration network $R$ and the translation network $T$ simultaneously, while encouraging $T$ to be geometry preserving. This ensures that the two networks are task-specific --- $T$ performs only a photo-metric mapping, while $R$ learns the geometric transformation required for the registration task. In our work, we use the concepts of generative adversarial networks (GAN~\cite{GAN, CGAN}) to train $T$ and $R$. We show that the adversarial training is not only necessary for the translation task (as shown in previous works~\cite{pix2pix2016}), but is also necessary to produce smooth and accurate spatial transformation. We evaluate our method on real commercial data, and demonstrate its strength with a series of studies.

The main contributions of our work are:
\begin{itemize}
    \item An unsupervised method for multi-modal image registration.
    \item A geometry preserving translation network that allows the application of mono-modality metrics in multi-modal registration.
    \item A training scheme that encourages a generator to be geometry preserving.
\end{itemize}{}

\section{Related Works}


To deal with the photo-metric difference between modalities, unsupervised multi-modal approaches are forced to find the correlation between the different domains and use it to guide their learning process. In \cite{elastic_registration_nostn_multi_justgan} a vanilla CycleGAN architecture is used to regularize a deformation mapping. This is achieved by training a discriminator network to distinguish between deformed and real images. To align a pair of images the entire network needs to be trained in a single pass. Training this network on a large dataset will encourage the deformation mapping to become an identity mapping. This is because the discriminator is given only the real and deformed images. Furthermore the authors use multiple cross-modality similarity metrics including SSIM, NCC and NMI which are limited by the compatibility of the specific modalities used. In contrast, our method learns from a large dataset and bypasses the need for cross-modality similarity metrics.


Wang \etal~\cite{unsupervised_stn_gan_multi_1_jointloss} attempt to bypass the need for domain translation by learning an Encoder-Decoder module to create  modality-independent features. The features are fed to an STN to learn affine and non-rigid transformations. The authors train their network using a simple similarity measure (MSE) which maintains local similarity, but does not enforce global fidelity.

At the other extreme, \cite{unsupervised_stn_gan_multi_3_justgan} rely entirely on an adversarial loss function. They train a regular U-Net based STN by giving the resultant registered images to a discriminator network and using its feedback as the STN's loss function. By relying solely on the discriminator network for guiding the training, they lose the ability to enforce local coherence between the registered and target images.

Closest to our work, \cite{unsupervised_gan_multi_2_nostn_justgan} combines an adversarial loss with similarity measurements in an effort to register the images properly while concentrating on maintaining local geometric properties. They encode the inputs into two separate embedding, one for shape and one for content information, and train a registration network on these disentangled embedding. This method relies on learned disentanglement, which introduces inconsistencies on the local level. Our method directly enforces the similarity in the image space, which leads to a reliable local signal.

\begin{figure*}[t]
    \centering
    \begin{subfigure}[m]{0.25\textwidth}
    \centering
    \includegraphics[width=0.9\textwidth]{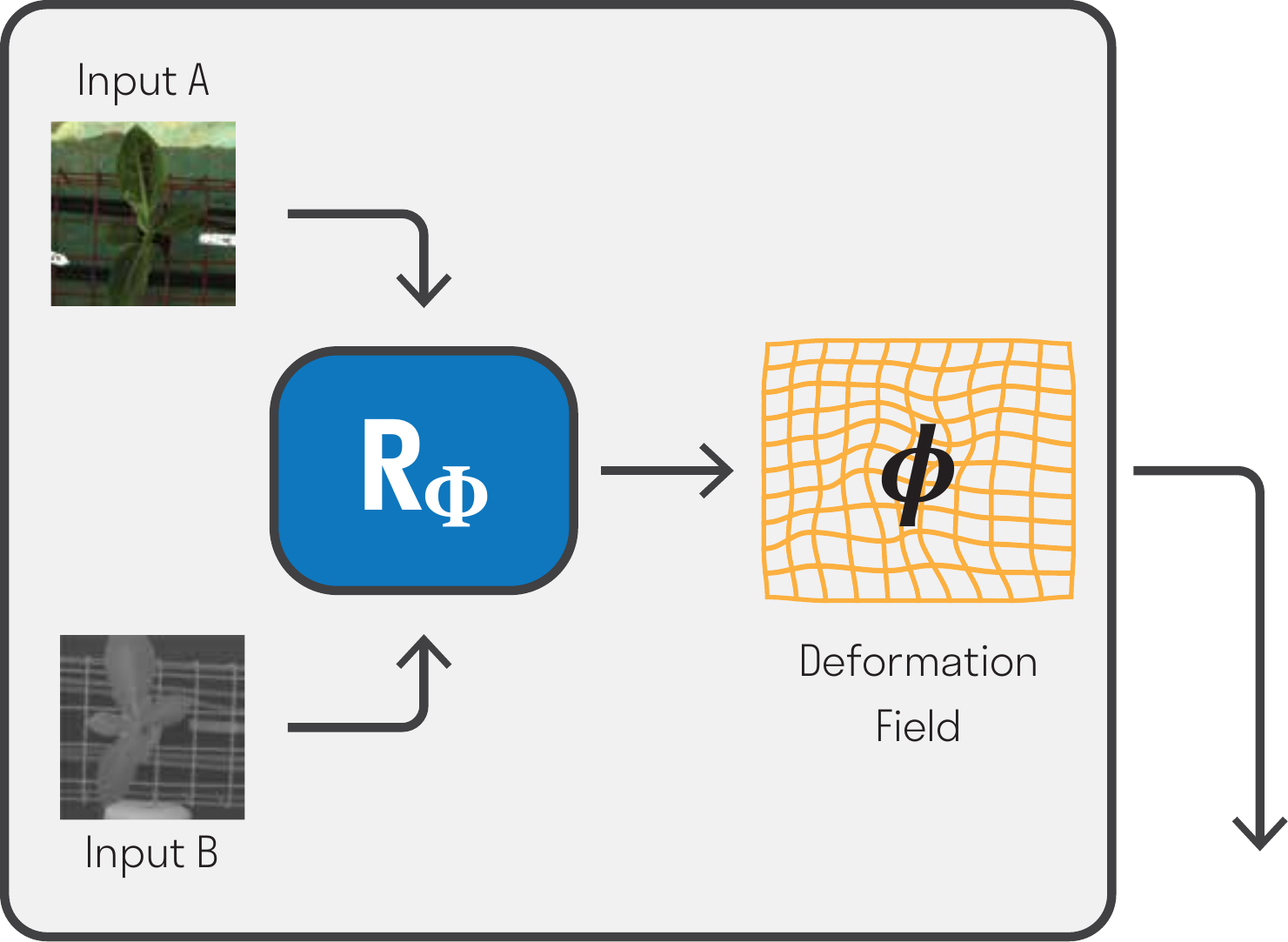}
    \caption{Deformation Field Generation}
    \label{fig:deform_gen}
    \end{subfigure}
    \begin{subfigure}[m]{0.35\textwidth}
    \centering
    \includegraphics[width=0.9\textwidth]{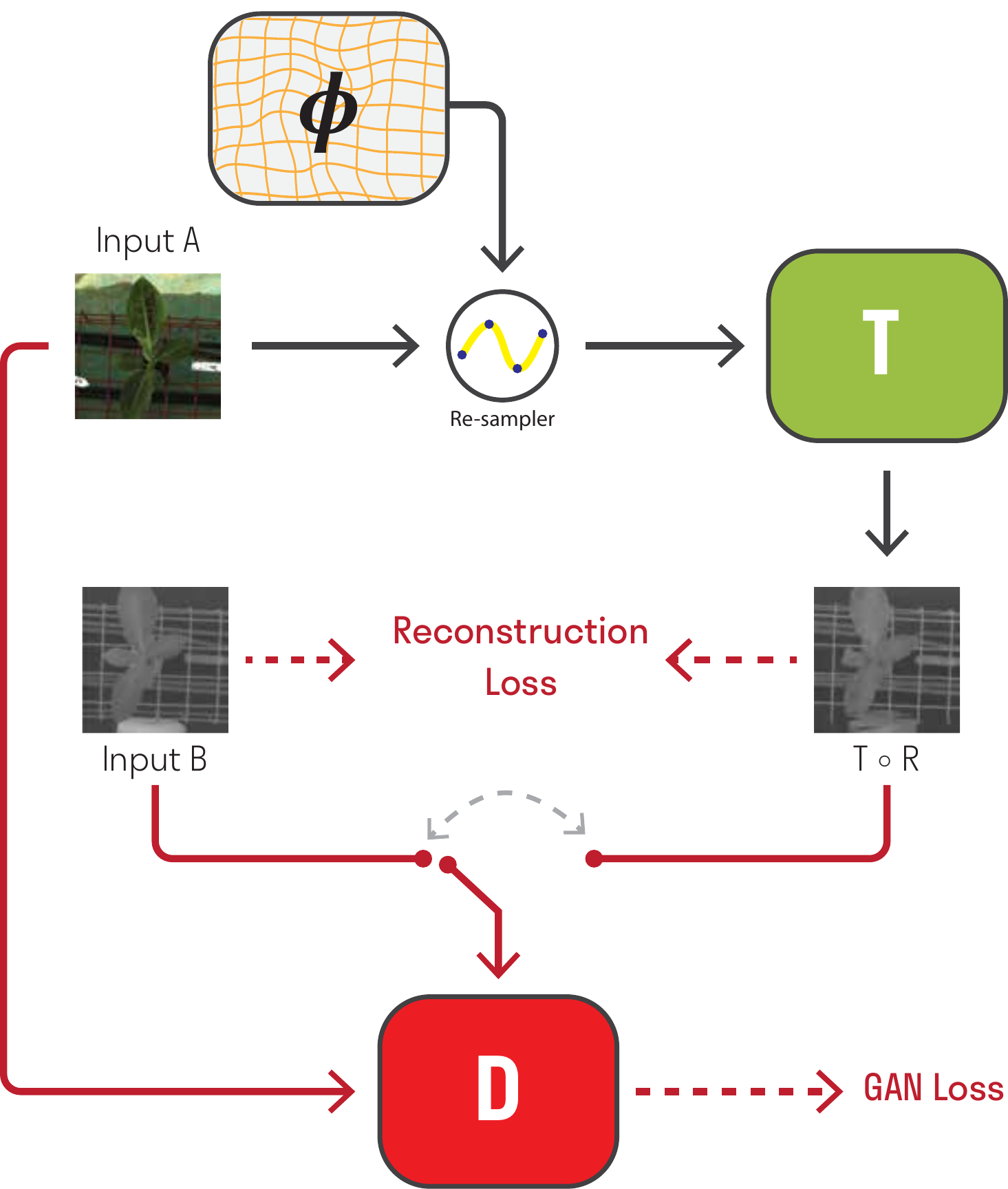}
    \caption{Register First}
    \label{fig:register_first}
    \end{subfigure}
    \begin{subfigure}[m]{0.35\textwidth}
    \centering
    \includegraphics[width=0.9\textwidth]{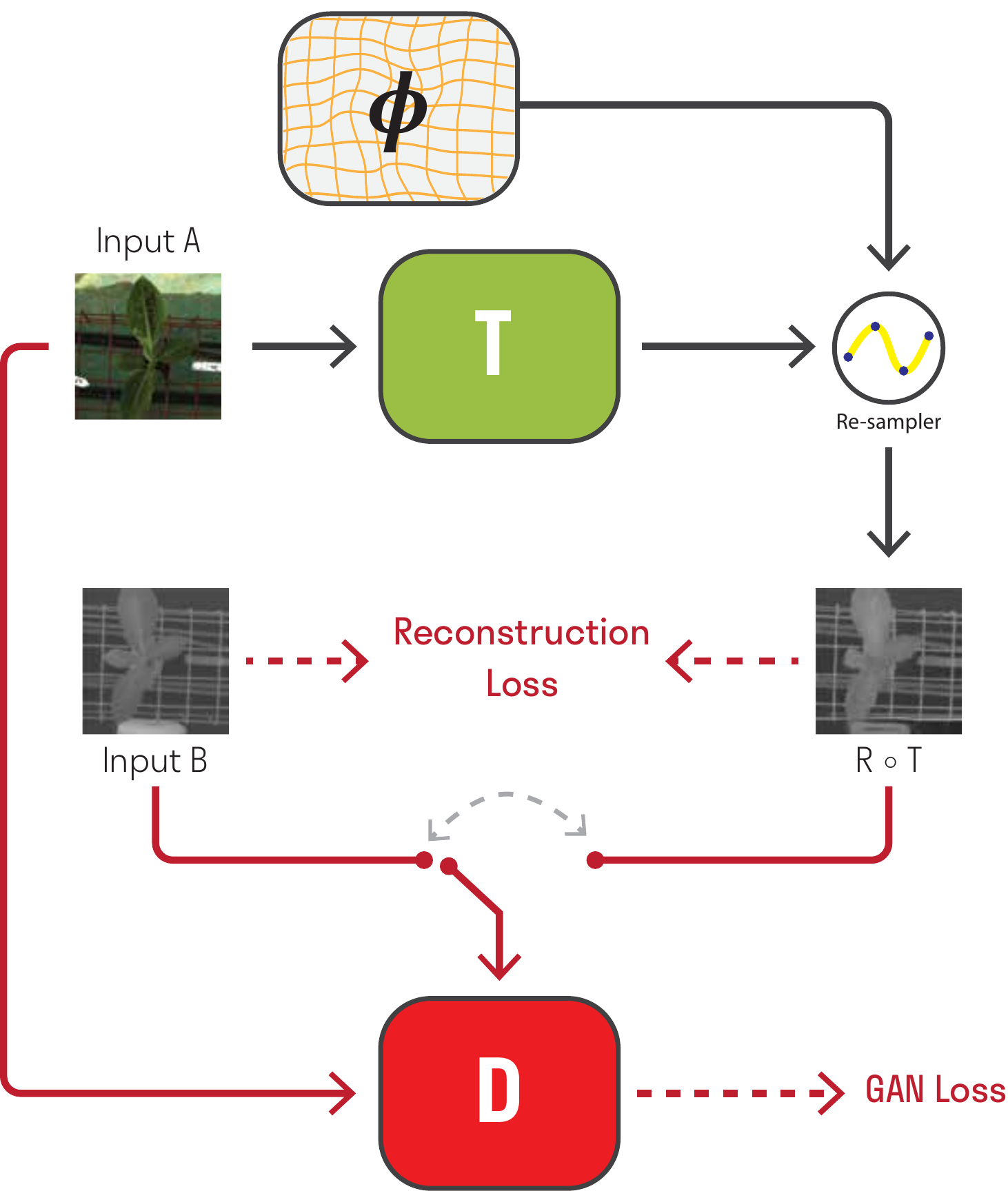}
    \caption{Translate First}
    \label{fig:translate_first}
    \end{subfigure}

    \caption{\textbf{Training Flow Overview.} We train two components: (i) a spatial transformation network (STN) $R = (R_{\Phi}, R_S)$ and (ii) an image-to-image translation network $T$. The two networks $R$ and $T$ are jointly trained via two different training flows. The two training flows are simultaneously carried-out in each training step. In the first flow, \emph{(b) Register First}, the input image $I_a$ is deformed using $\phi$, a deformation field generated by $R_{\Phi}$, and is then fed to $T$ to map the image onto domain B. The second flow, \emph{(c) Translate First}, is similar with the exception that $\phi$ is used to transform the  \emph{translated} source image. In both cases, the same deformation field $\phi$ is used.
    }
    
    
    \label{fig:training_overview}
\end{figure*}

\section{Overview}

Our core idea is to \textit{learn} the translation between the two modalities, rather than using a cross-modality metric.
This novel approach is illustrated in Figure \ref{fig:NeMAR}.
The spatially transformed image is translated by a learnable network. The translated image can then  be compared to the original source image using a simple uni-modality metric, bypassing the need to use a cross-modality metric. The advantage of using a learnable translation network is that it generalizes and adapts to any pairs of given modalities.

Our registration network consists of two components: (i) a spatial transformation network $R = (R_{\Phi}, R_S)$ and (ii) an image-to-image translation network $T$. The two components are trained simultaneously using two training flows as depicted in Figure~\ref{fig:training_overview}. The spatial transformation network takes the two input images and yields a deformation field $\phi$. The field is the then applied either before $T$ (Figure~\ref{fig:register_first}) or after  it (Figure~\ref{fig:translate_first}). Specifically, the field is generated using a network $R_\Phi$ and is used by a re-sampling layer $R_{S}$ to get the transformed image, namely $R_{S}(T(a),\phi)$ and $T(R_{S}(a,\phi))$. We will elaborate on these two training schemes in Section~\ref{sec:geo_preserving_g}. The key is, as we shall show, that such two-flow training encourages $T$ to be geometry preserving, which implies that all the geometry transformation is encoded in $R_{\Phi}$.  

Once trained, only the spatial transformation network $R$ is used in test time. The network takes two images $I_a$ and $I_b$ representing the same scene, captured from slightly different viewpoint, in two different modalities, $A$ and $B$, respectively, and aligns $I_a$ with $I_b$.

\section{Method}
Our goal is to learn a non-rigid spatial transformation which aligns two images from different domains. Let $\mathcal{A} \subset \mathbb{R}^{H_A\times W_A \times C_A}$ and $\mathcal{B} \subset \mathbb{R}^{H_B\times W_B \times C_B}$ be two paired image domains, where $H_D,W_D,C_D$ are the height, width, and number of channels for domain $\mathcal{D}$, respectively. Pairing means that for each image $I_a \in \mathcal{A}$ there exists a unique image $I_b \in \mathcal{B}$ representing the same scene, as acquired by the different respective sensors. Note that the pairing assumption is a common and reasonable one, since more often than not registration-base applications involve taking an image of the same scene from both modality sensors (e.g satellite images). Throughout this section, we let $I_a\in \mathcal{A}$ and $I_b \in \mathcal{B}$ be a pair of two images such that $I_a$ needs to be aligned with $I_b$.

To achieve this alignment, we train three learnable components: (i) a registration network $R$, (ii) a translation network $T$ and (iii) a discriminator $D$. The three networks are trained using an adversarial model \cite{GAN, CGAN}, where $R$ and $T$ are jointly trained to outwit $D$.  Below, we describe the design and  objectives of each network.

\subsection{Registration Network}

Our registration network ($R=(R_{\Phi}, R_{S})$) is a spatial transformation network (STN) composed of a fully-convolutional network $R_{\Phi}$ and a re-sampler layer $R_S$. The transformation we apply is a non-linear dense deformation - allowing non-uniform mapping between the images and hence gives accurate results. Next we give an in-depth description about each component.

\textbf{$\mathbf{R_{\Phi}}$ - Deformation Field Generator:} The network takes two input images, $I_a$ and $I_b$, and produces a deformation field $\phi=R(I_a, I_b)$ describing how to non-rigidly align $I_a$ to $I_b$. The field is an $H_A\times W_A$ matrix of $2$-dimensional vectors, indicating the deformation direction for each pixel $(i,j)$ in the input image $I_a$.

\textbf{$\mathbf{R_S}$ - Re-sampling Layer:} This layer receives the deformation field $\phi$, produced by $R_{\Phi}$, and applies it on a source image $I_s$. Here, the source image is not necessarily $I_a$ and it could be from either domains - $\mathcal{A}$ or $\mathcal{B}$. Specifically, the value of the transformed image $R_S(I_s, \phi)$ at pixel $\mathbf{v}=(i,j)$ is given by Equation~\ref{eq:deformed_image}:

\begin{equation}
\label{eq:deformed_image}
R_{S}(I_s, \phi)[v] = I_s \left[ \mathbf{v} + \phi(\mathbf{v}) \right],
\end{equation}
where $\phi(\mathbf{v}) = (\Delta y, \Delta x)$ is the deformation generated by $R_{\Phi}$ at pixel $\mathbf{v}=(i,j)$, in the $x$ and $y$-directions, respectively.

To avoid overly distorting the deformed image $R_S(I_s,\phi)$ we restrict $R_{\Phi}$ from producing non-smooth deformations. We adapt a common regularization term that is used to produce smooth deformations. In particular, the regularization loss will encourage neighboring pixels to have similar deformations. Formally, we seek to have small values of the first order gradients of $\phi$, hence the loss at pixel $\mathbf{v}=(i,j)$ is then given by:
\begin{equation}
\label{eq:smooth}
\mathcal{L}_{smooth}(\phi, \mathbf{v}) = \sum_{\mathbf{u} \in N(\mathbf{v})} B(\mathbf{u},\mathbf{v}) \norm{ \phi(\mathbf{u}) - \phi(\mathbf{v})},
\end{equation}
where $N(\mathbf{v})$ is a set of neighbors of the pixel $\mathbf{v}$, and $B\left(\mathbf{u}, \mathbf{v} \right)$ is a bilateral filter~\cite{Bilateral} used to reduce over-smoothing. Let $O_s = R_S(I_s,\phi)$ to be the deformed image produced by $R_S$ on input $I_s$, then the bilateral filter is given by:
\begin{equation}
\label{eq:bilateral}
B(\mathbf{u},\mathbf{v}) = e^{-\alpha \cdot \norm{O_s[\mathbf{u}] - O_s[\mathbf{v}]}}.
\end{equation}

There are two important notes about the bilateral filter $B$ in Equation~\ref{eq:bilateral}. First, the bilateral filtering is with respect to the transformed image $O_s$ (at each forward pass), and secondly, the term $B\left(I_s, \mathbf{u}, \mathbf{v}\right)$ is a treated as constant (at each backward pass). The latter is important to avoid $R_{\Phi}$ alternating pixel values so that $B(u,v) \approx 0$ (e.g., it could change pixels so that $\norm{O_s[\mathbf{u}] - O_s[\mathbf{v}]}$ is relatively large), while the former allows better exploration of the solution space.

In our experiments we look at the $3 \times 3$ neighborhood of $\mathbf{v}$, and set $\alpha=1$. The overall smoothness loss of the network $R$, denoted by $\mathcal{L}_{smooth}\left(R\right)$,  is the mean value over all pixels $\mathbf{v} \in \left\{1, \dots{}, H_A \right\} \times \left\{1, \dots{}, W_A \right\}$.

\subsection{Geometric Preserving Translation Network}
\label{sec:geo_preserving_g}

A key challenge of our work is to train the image-to-image translation network $T$  to be \textbf{geometric preserving}. If $T$ is geometric preserving, it implies that it only performs photo-metric mapping, and as a consequence the registration task is performed solely by the registration network $R$. However, during our experiments, we observed that $T$ tends to generate fake images that are spatially aligned with the ground truth image, regardless of $R$'s accuracy.

To avoid this, we could restrict $T$ from performing any spatial alignment by reducing its capacity (number of layers). While we did observe that reducing $T$'s capacity does improve our registration network's performance, it still limits the registration network from doing all the registration task (See supplementary materials). 

To implicitly encourage $T$ to be geometric preserving we require that $T$ and $R$ are commutative, i.e., $T \circ R = R \circ T$. In the following we formally define both $T \circ R$ and $R \circ T$:

\paragraph{Translation First - $\mathbf{\left(R \circ T \right) \left( I_a, I_b\right)}$:} This mapping first apply an image-to-image translation on $I_a$ and then a spatial transformation on the translated image. Specifically, the final image is obtained by first applying $T$ on $I_a$, which generates a fake sample $O_{T} = T(I_a)$. Then we apply our spatial transformation network  $R$ on $O_T$ and get the final output:
$$O_{RT} = R_{S}\left(O_T, \phi\right) = R\left(T\left(I_a\right),R_{\Phi}\left(I_a, I_b\right)\right).$$

\paragraph{Register First - $\mathbf{\left(T \circ R \right) \left( I_a, I_b\right)}$} in this composition, we first apply spatial transformation on $I_a$ and obtain a deformed image $O_R = R(I_a, \phi)$. Then, we translate $O_R$ to domain $\mathcal{B}$ using our translation network $T$: 
$$O_{TR} = T\left(R_{S}\left(I_a, \phi\right)\right) = T\left(R_{S}\left(I_a, R_{\Phi}\left(I_a, I_b \right)\right)\right).$$

Note that in both compositions (i.e., $T\circ R$ and $R \circ T$), the deformation field, used by the re-sampler $R_S$, is given by $R_{\Phi}\left(I_a, I_b\right)$. The only difference is in the source image from which we re-sample the deformed image. Throughout this section, we refer to $O_{RT}$ and $O_{TR}$ as the outputs of $R\circ T$ and $T \circ R$, respectively.

\subsection{Training Losses}
To train $R$ and $T$ to generate fake samples that are similar to those in domain $\mathcal{B}$, we use an $L1$-reconstruction loss:
\begin{equation}
\label{eq:reconstruction_loss}
\mathcal{L}_{recon}(T,R) = \norm{ O_{RT} - I_b}_1 + \norm{ O_{TR} - I_b}_1 
\end{equation}
where minimizing the above implies that $T \circ R \approx R \circ T$. 

We use conditional GAN (cGAN)\cite{CGAN} as our adversarial loss for training $D$, $T$ and $R$. The objective of the adversarial network $D$ is to discriminate between real and fake samples, while $T$ and $R$ are jointly trained to fool the discriminator. The cGAN loss for $T\circ R$ and $R \circ T$ is formulated below:
\begin{equation}
\label{eq:cgan_gr}
\begin{split}
    \mathcal{L}_{cGAN}\left(T,R,D\right)  = & \mathbb{E}\left[\log\left(D\left(I_b, I_a\right)\right) \right] \\
    & +\mathbb{E}\left[\log(1-D(O_{RT}, I_a)) \right] \\
    & +\mathbb{E}\left[\log(1-D(O_{TR}, I_a)) \right],
\end{split}
\end{equation}

The total objective is given by:

\begin{equation}
\label{eq:total_loss}
\begin{split}
\mathcal{L}(T,R) = & \underset{D}{\arg\max} \mathcal{L}_{cGAN}(T,D,R) \\ 
& + \lambda_{R} \cdot \mathcal{L}_{recon}(T,R) + \lambda_{S} \cdot \mathcal{L}_{smooth}(R) ,
\end{split}
\end{equation}
where we are opt to find $T^{*}$ and $R^{*}$ such that $T^{*}, R^{*} = \underset{R,T}{\arg \min} \mathcal{L} (T, R)$. Furthermore, in our experiments, we set $\lambda_R = 100$ and $\lambda_S = 200$.

\subsection{Implementation Details}
Our code is implemented using PyTorch 1.1.0~\cite{paszke2017automatic} and is based on the framework and implementation of Pix2Pix~\cite{pix2pix2016}, CycleGAN~\cite{CycleGAN2017} and BiCycleGAN~\cite{zhu2017toward}. The network $T$ is an encoder-decoder network with residual connections~\cite{2015DeepRL}, which is adapted from the implementation in~\cite{Johnson2016Perceptual}. The registration network is U-NET based~\cite{UNET} with residual connections in the encoder-path and the output path. In all residual connections, we use Instance Normalization Layer~\cite{Ulyanov2016InstanceNT}. All networks were initialized by the Kaiming~\cite{He2015} initialization method. Full details about the architectures is provided in the supplementary material. 

The experiments were conducted on single GeForce RTX 2080 Ti. We use Adam Optimizer~\cite{Kingma2014AdamAM} on a mini-batch of size 12 with parameters $lr=1 \times e^{-4}$, $\beta_1 = 0.5$ and $\beta_2=0.999$. We train our model for 200 epochs, and activate linear learning rate decay after 100 epochs.



%


\section{Experimental Results}

In the following section we evaluate our approach and explore the interactions between $R$, $T$ and the different loss terms we use.

All our experiments were conducted on a commercial dataset, which contains a collection of images of banana plants with different growing conditions and phenotype. The dataset contains 6100 image frames, where each frame consist of an RGB image, IR Image and Depth Image. The colored images are a 24bit Color Bitmap captured from a high-resolution sensor. The IR images are a 16bit gray-scale image, captured from a long-wave infrared (LWIR) sensor. Finally, the depth images were captured by Intel Real-Sense depth camera. The three sensors were calibrated, and an initial registration was applied based on affine transformation estimation via depth and controlled lab measurements. The misalignment in the dataset is due to depth variation within different objects in the scene, which the initial registration fails to handle. We split the dataset into training and test samples, where the test image were sampled with probability $p=0.1$. 

\subsection{Evaluation}
\begin{figure}
    \centering
    \includegraphics[width=0.46\textwidth]{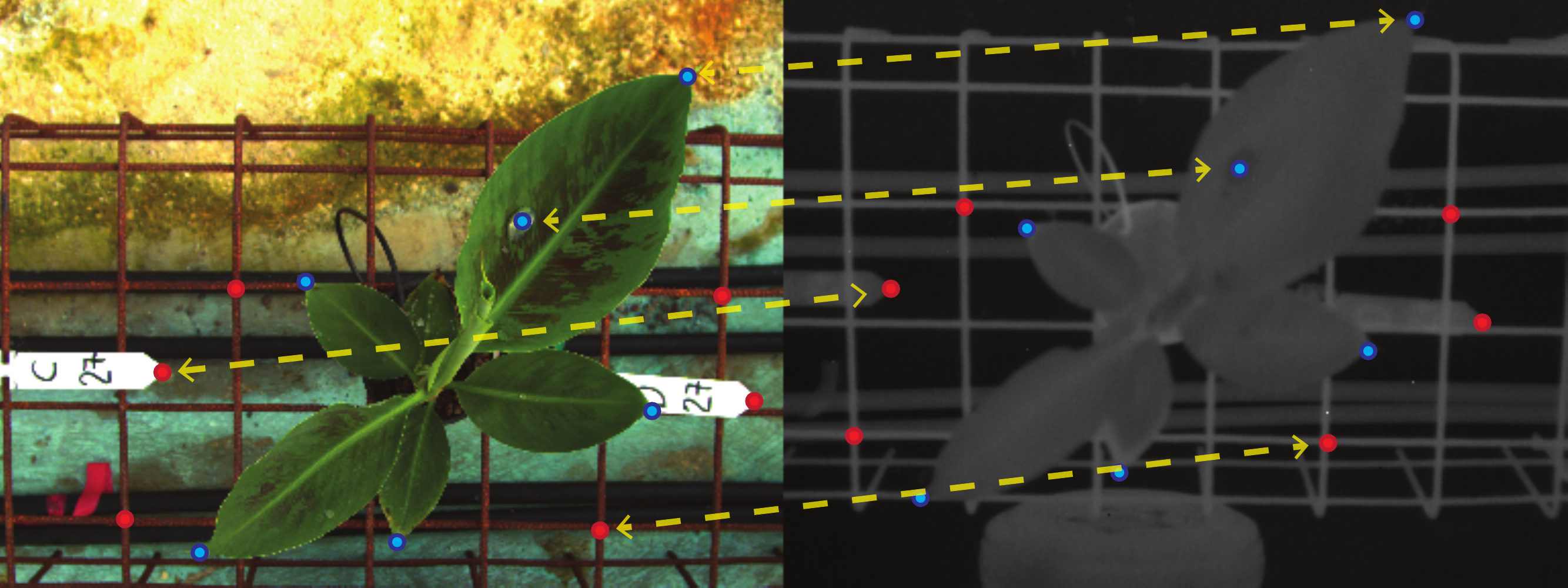}
    \caption{\textbf{Annotation sample.} An example of image demonstrating our annotations. We pick points from both the source image $I_a$ (RGB image on the left) and the target image $I_b$ (Thermal image on the right). The blue points are on salient objects and the red points are general points from the scene. We added several arrows to illustrate some matching points. Further, the geometry of each point is with respect to its corresponding image.}
    \label{fig:annotation_example}
\end{figure}{}
\textbf{Registration Accuracy Metric.} We manually annotated 100 random pairs of test images. We tagged 10-15 pairs of point landmarks on the source and target images which are notable and expected to match in the registration process (See Figure~\ref{fig:annotation_example}).  Given a pair of test images, $I_a$ and $I_b$, with a set of tagged pairs ${(x_{a_i}, y_{a_i}), (x_{b_i}, y_{b_i})}_{i\in[N]}$, denote $(x'_{a_i} ,y'_{a_i})$ as deformed sample points $(x_{a_i}, y_{a_i})$. The accuracy of the registration network $R$ is simply the average Euclidean distance between the target points and the deformed source points:

\begin{equation}
    \label{eq:reg_acc}
    acc = \frac{1}{N} \sum_{i\in \left[N\right]} \sqrt{(x'_{a_i}-x_{b_i})^2 + (y'_{a_i}-y_{b_i})^2}.
\end{equation}

Furthermore, we used two type of annotations. The first type of annotation is located over salient objects in the scene (the blue points in Figure~\ref{fig:annotation_example}). This is important because in most cases, down-stream tasks are affected mainly by the alignment of the main object in the scene in both modalities. The second annotation is performed by picking landmark points from all objects across the scene.

\begin{table}
    \centering
    \begin{center}
        \begin{tabular}{||l c c||} 
        \hline
        Method & Salient Objects & Full Scene \\ [0.5ex] 
        \hline\hline
        Unregistered & 30.3 & 35.45 \\
         \hline
        CG~\cite{CycleGAN2017} + SIFT~\cite{SIFT} & 17.9 & 34.74  \\ 
        \hline
        $R$ + SSIM On Edges & 26.12 & 28.41 \\
        \hline
        $R$ + NCC On Edges & 16.78 & 27.41  \\
        \hline
        $R$ + NCC &  15.8 & 29.91 \\
        \hline
        $R$ + $T$ \textbf{(Ours)} & 6.27 & 6.93  \\ 
        \hline    
        \end{tabular}
        \end{center}
    \caption{\textbf{Registration Accuracy results.} Registration accuracy for various similarity measurements. Unregistered (first row), represents the misalignment in the dataset. CG + SIFT is training a CycleGAN and using SIFT features on the generated images. Finally, we also show the registration accuracy of training our registration network with different loss terms.  }
    \label{table:numerical_comparison_with_metrics}
\end{table}

\textbf{Quantitative Evaluation.} Due to limited access to source code and datasets of related works, we conduct several experiments that demonstrate the power of our method with respect to different aspects of previous works. As the crux of our work is the alleviation of the need for cross-modality similarity measures, we trained our network with commonly used cross-modality measures. In Table~\ref{table:numerical_comparison_with_metrics} we show the registration accuracy of our registration network $R$ when trained with different loss terms. Specifically, we used Normalized Cross Correlation as it is frequently used in unsupervised multi-modal registration methods. Furthermore, we trained our network with Structural Similarity Index Metric (SSIM) on edges detected by Canny edge-detector~\cite{Canny1986} from both the deformed and target image. Finally, we attempt to train our registration network $R$ by maximizing the Normalized Cross Correlation between the edges of the deformed and the target image.  As can be seen from Table~\ref{table:numerical_comparison_with_metrics}, training the registration network $R$ using prescribed cross-modality similarity measures do not perform well. Further, using these $NCC$ produces noisy results, while using SSIM gives smooth but less accurate registration (see supplemental materials).

Furthermore, we also tried using traditional descriptors such as SIFT~\cite{SIFT} in order to match corresponding key points from the source and target image. We use these key points to register the source and target images by estimating the transformation parameters between them. However, these descriptors are not designed for multi-modal data, and hence they fail badly to be used on our dataset.

Instead, we train a CycleGAN \cite{CycleGAN2017} network to translate between the two modalities at hand, without any supervision to match the ground truth. CycleGAN, like other unsupervised image-to-image translation networks, is not trained to generate images matching  ground truth samples, thus, geometric transformation is not explicitly required from the translation network. Once trained, we use one of the generators in the CycleGAN, the one that maps between domain $\mathcal{A}$ to domain $\mathcal{B}$ to translate the input image $I_a$ onto modality $\mathcal{B}$. Assuming this generator is both geometry preserving and translates well between the modalities, it is expected that it also match well between feature of the fake sample and the target image. Thus, we extracted SIFT descriptors from the generated images by the CycleGAN translation network, and extracted SIFT features from the target image $I_b$. We then matched these features and estimated the needed spatial registration. The registration accuracy using this method is significantly better than directly using SIFT~\cite{SIFT} features on the input image $I_a$. The results are shown in Table~\ref{table:numerical_comparison_with_metrics}. Further visual results and details demonstrating this method are provided in the supplementary material.

\textbf{Qualitative Evaluation.}
\begin{figure*}
    \centering
    \begin{subfigure}{\textwidth}
        \centering
        
    \begin{subfigure}{0.158\textwidth}
    \includegraphics[width=\textwidth]{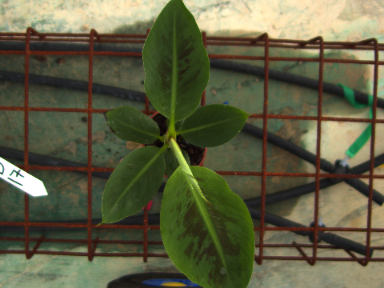}
    \includegraphics[width=\textwidth]{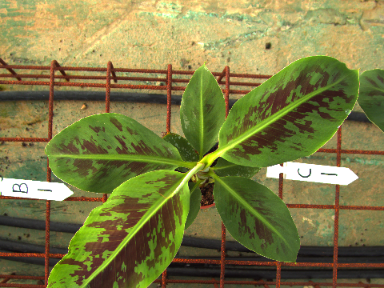}
    \includegraphics[width=\textwidth]{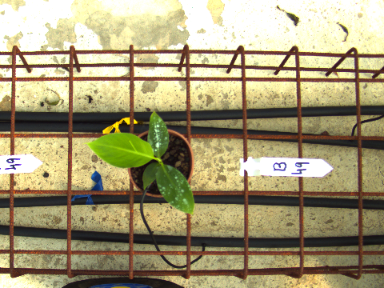}
    \caption*{Input A}

    \end{subfigure}
    \begin{subfigure}{0.158\textwidth}
    \includegraphics[width=\textwidth]{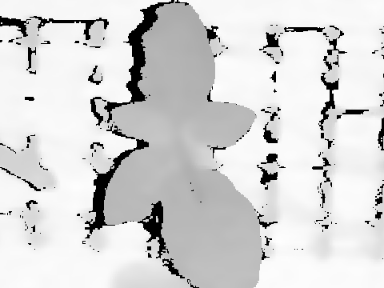}
    \includegraphics[width=\textwidth]{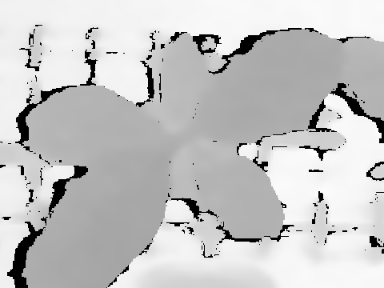}
    \includegraphics[width=\textwidth]{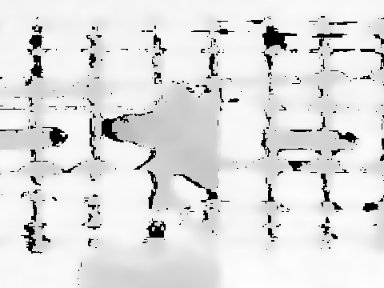}
    \caption*{Input B}
    \end{subfigure}
    \begin{subfigure}{0.158\textwidth}
    \includegraphics[width=\textwidth]{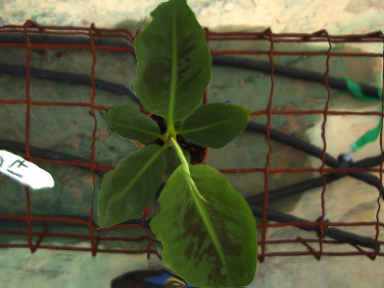}
    \includegraphics[width=\textwidth]{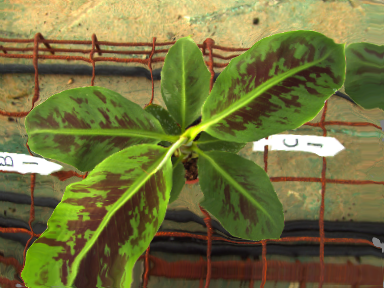}
    \includegraphics[width=\textwidth]{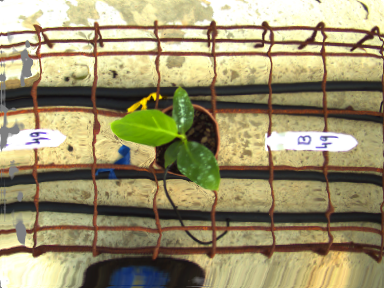}
    \caption*{Registered}
    \end{subfigure}
    \begin{subfigure}{0.158\textwidth}
    \includegraphics[width=\textwidth]{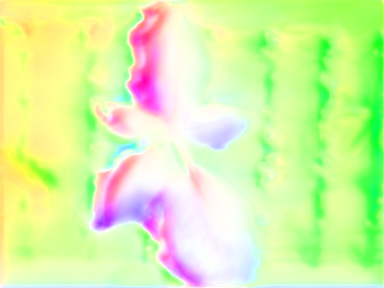}
    \includegraphics[width=\textwidth]{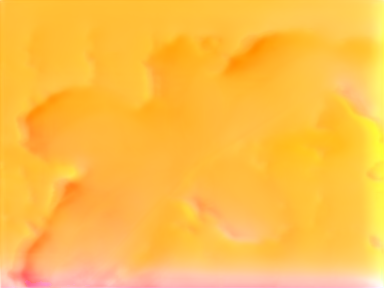}
    \includegraphics[width=\textwidth]{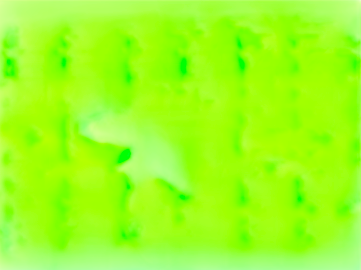}
    \caption*{Deform. Field}
    \end{subfigure}
    \begin{subfigure}{0.158\textwidth}
    \includegraphics[width=\textwidth]{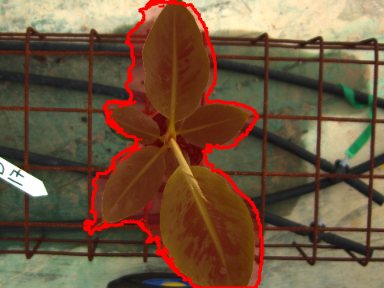}
    \includegraphics[width=\textwidth]{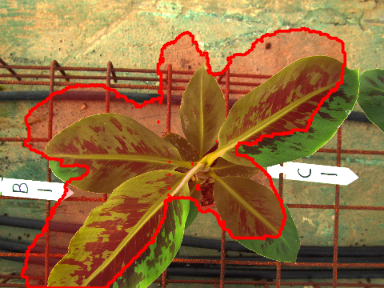}
    \includegraphics[width=\textwidth]{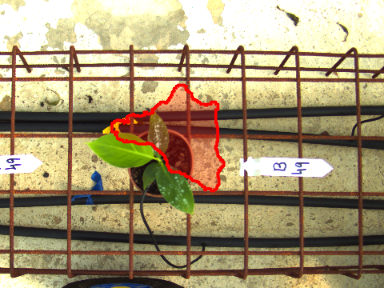}
    \caption*{Before}
    \end{subfigure}
    \begin{subfigure}{0.158\textwidth}
    \includegraphics[width=\textwidth]{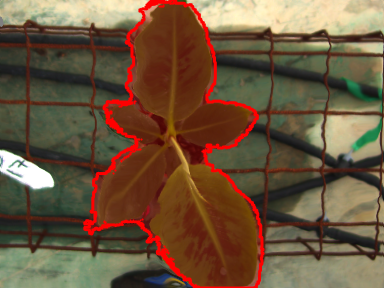}
    \includegraphics[width=\textwidth]{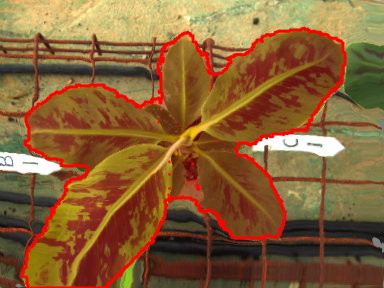}
    \includegraphics[width=\textwidth]{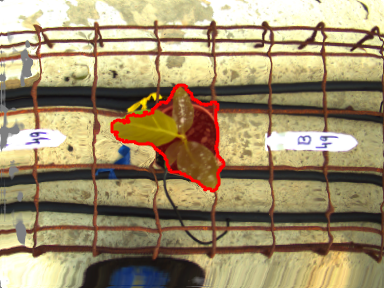}
    \caption*{After}
    \end{subfigure}
    \caption{Image registration between RGB and Depth modalities.}
    \label{fig:perceptual_rgb_to_depth}
    \end{subfigure}
    
    \vspace{0.025\textwidth}

    \begin{subfigure}{\textwidth}
    \centering
    \begin{subfigure}{0.158\textwidth}
    \includegraphics[width=\textwidth]{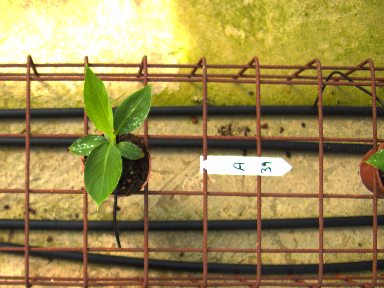}
    \includegraphics[width=\textwidth]{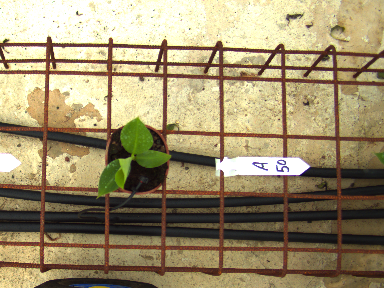}
    \includegraphics[width=\textwidth]{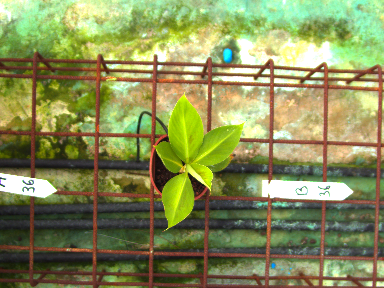}
    \caption*{Input A}
    \end{subfigure}
    \begin{subfigure}{0.158\textwidth}
    \includegraphics[width=\textwidth]{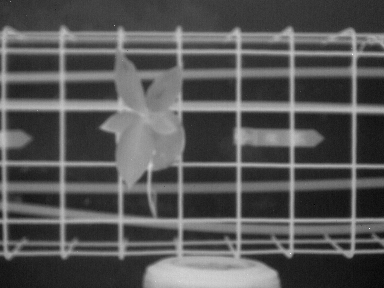}
    \includegraphics[width=\textwidth]{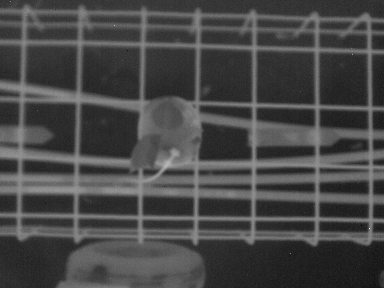}
    \includegraphics[width=\textwidth]{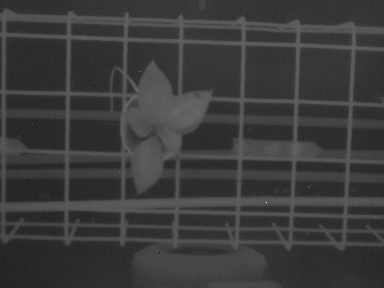}
    \caption*{Input B}
    \end{subfigure}
    \begin{subfigure}{0.158\textwidth}
    \includegraphics[width=\textwidth]{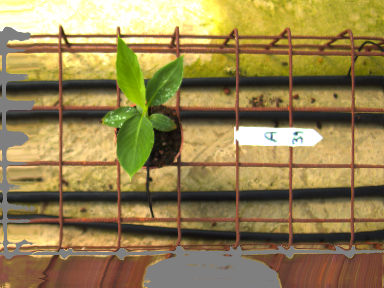}
    \includegraphics[width=\textwidth]{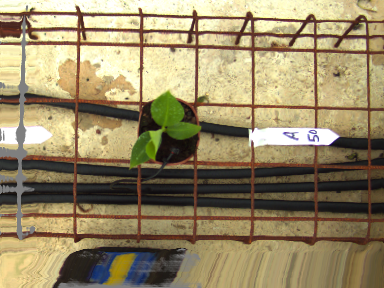}
    \includegraphics[width=\textwidth]{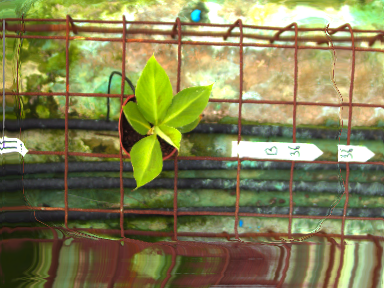}
    \caption*{Registered}
    \end{subfigure}
    \begin{subfigure}{0.158\textwidth}
    \includegraphics[width=\textwidth]{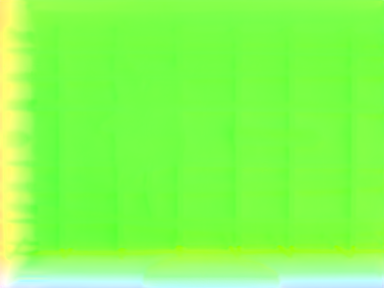}
    \includegraphics[width=\textwidth]{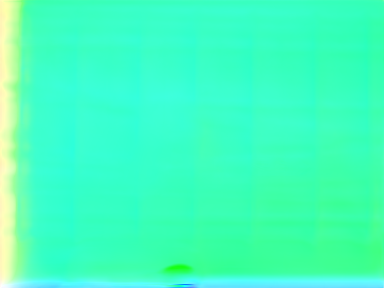}
    \includegraphics[width=\textwidth]{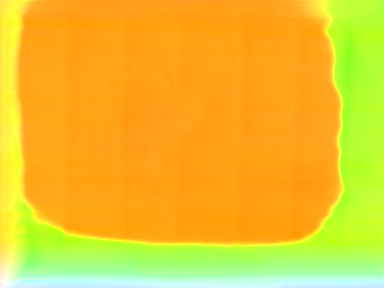}
    \caption*{Deform. Field}
    \end{subfigure}
    \begin{subfigure}{0.158\textwidth}
    \includegraphics[width=\textwidth]{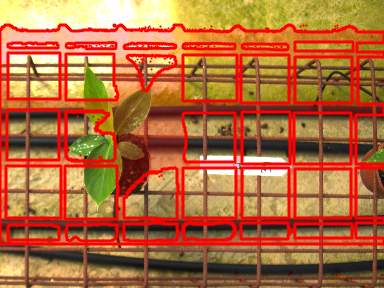}
    \includegraphics[width=\textwidth]{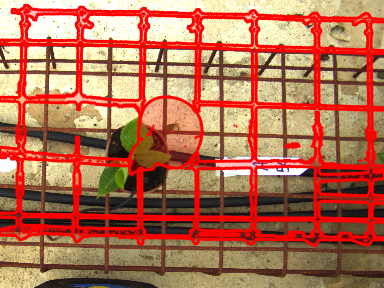}
    \includegraphics[width=\textwidth]{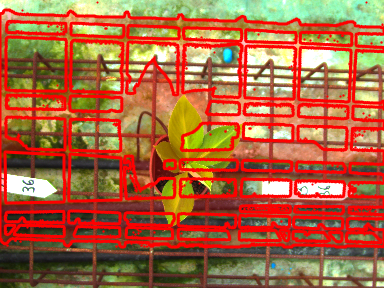}
    \caption*{Before}
    \end{subfigure}
    \begin{subfigure}{0.158\textwidth}
    \includegraphics[width=\textwidth]{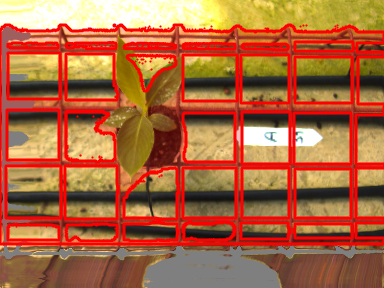}
    \includegraphics[width=\textwidth]{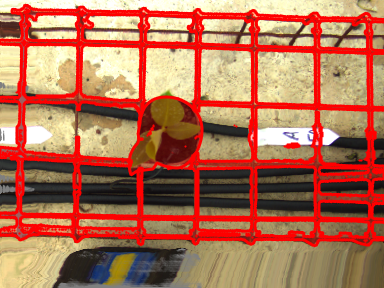}
    \includegraphics[width=\textwidth]{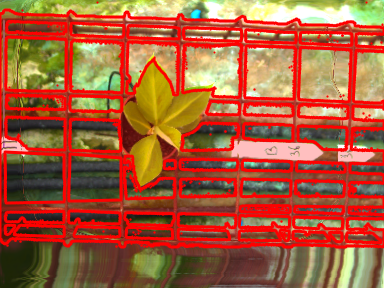}
    \caption*{After}
    \end{subfigure}
    \caption{Image registration between RGB and IR modalities.}
    \label{fig:perceptual_rgb_to_ir}
    \end{subfigure}
    
    \caption{\textbf{Qualitative Evaluation.} We show sample results on the registration between  two pairs of domains; (a) RGB to Depth registration and (b) RGB to IR registration. In the first two columns we show the corresponding images $I_a$ and $I_b$. The third column is the registered image, i.e the image $I_a$ after deformation. The deformation field (4th column) is visualized using the standard optical-flow visualization~\cite{OpticalFlowVisual}. Finally, we segment the salient object in $I_b$ and overlay it (with opacity 25\%) in the same spatial location onto the image before and after registration (last two columns).}
    \label{fig:qual_fig}
\end{figure*}
Figure~\ref{fig:qual_fig} shows that our registration network successfully aligns images from different pair of modalities and handles different alignment cases. For example, the banana leaves in the first raw in Figure~\ref{fig:perceptual_rgb_to_depth} are well-aligned in the two modalities. Our registration network maintains this alignment and only deforms the background for full alignment between the images. This can be seen from the deformation field visualization~\cite{OpticalFlowVisual}, where little deformation is applied on the banana plant, while most of the deformation is applied on the background. Furthermore, in the last row in Figure~\ref{fig:perceptual_rgb_to_depth}, there is little depth variation in the scene because the banana plant is small, hence a uniform deformation is applied across the entire image. To help measuring the alignment success, we overlay (with semi-transparency) the plant in image B on top of both image A before and after the registration. This means that the silhouette has the same spatial location in all images (the original image B, image A before and after the registration). Lastly, we achieve similar success in the registration between RGB and IR images (see Figure~\ref{fig:perceptual_rgb_to_ir}). 

It is worth mentioning that in some cases, the deformation field points to regions outside the source image. In those cases, we simply sample zero values. This happens because the the target image content (i.e., $I_b$) in these regions is not available in the source image (i.e., $I_a$). We provide more qualitative results in the supplemental materials.

\subsection{Ablation Study}
Next, we present a series of ablation studies that analyze the effectiveness of different aspects in our work. First, we show that training both compositions (i.e our presented two training flows) of $T$ and $R$ indeed encourages a geometric preserving translator $T$. Additionally, we analyze the impact of the different loss terms on the registration network's accuracy. We further show the effectiveness of the bilateral filtering, and that it indeed improves the registration accuracy. All experiments, unless otherwise stated, were conducted without the bilateral filtering.

\textbf{Geometric-Preserving Translation Network.}

\begin{figure*}
    \centering
    \begin{subfigure}{0.2375\textwidth}
    \centering
    \includegraphics[width=\textwidth]{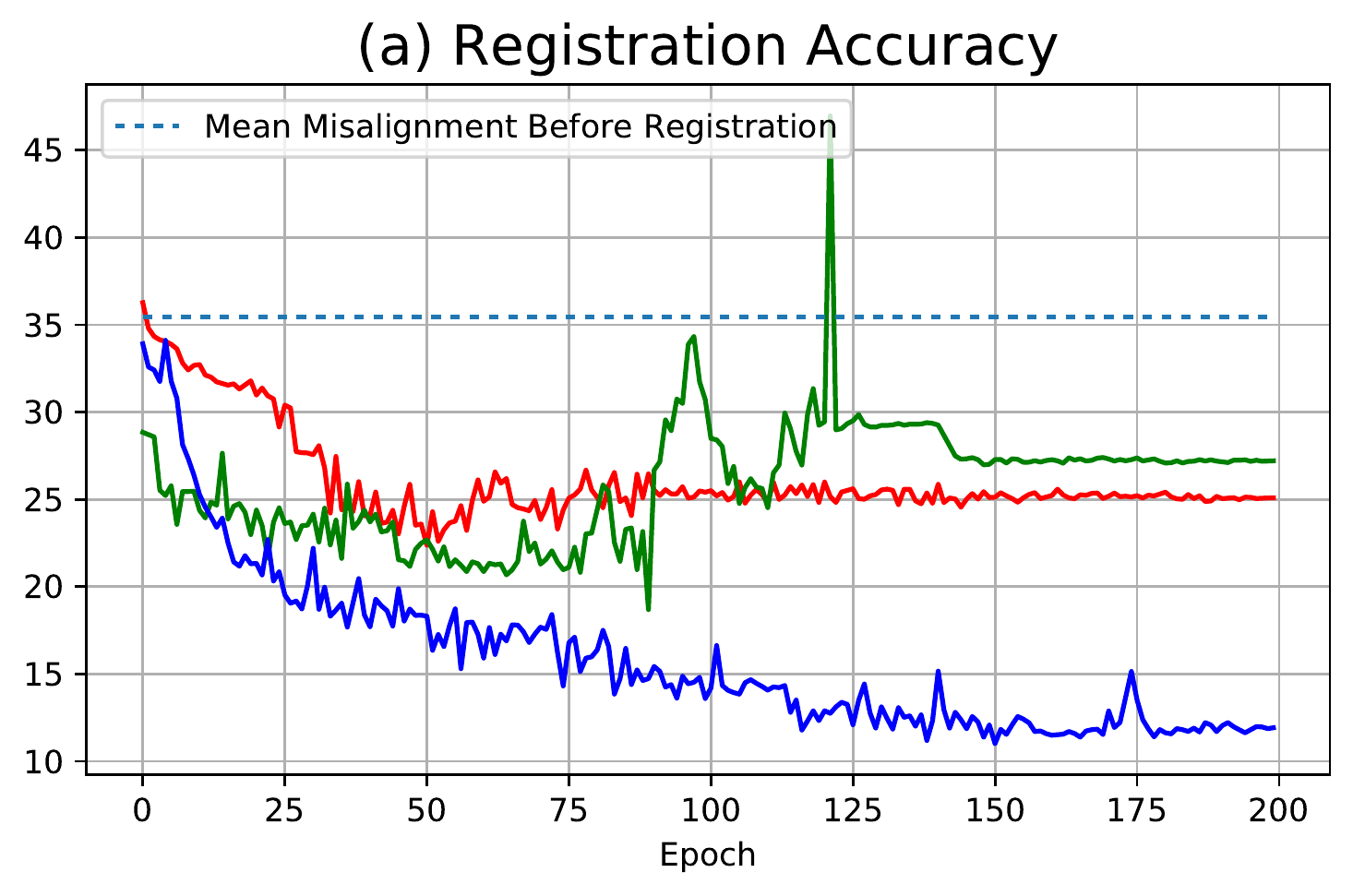}
    \end{subfigure}
    \begin{subfigure}{0.2375\textwidth}
    \centering
    \includegraphics[width=\textwidth]{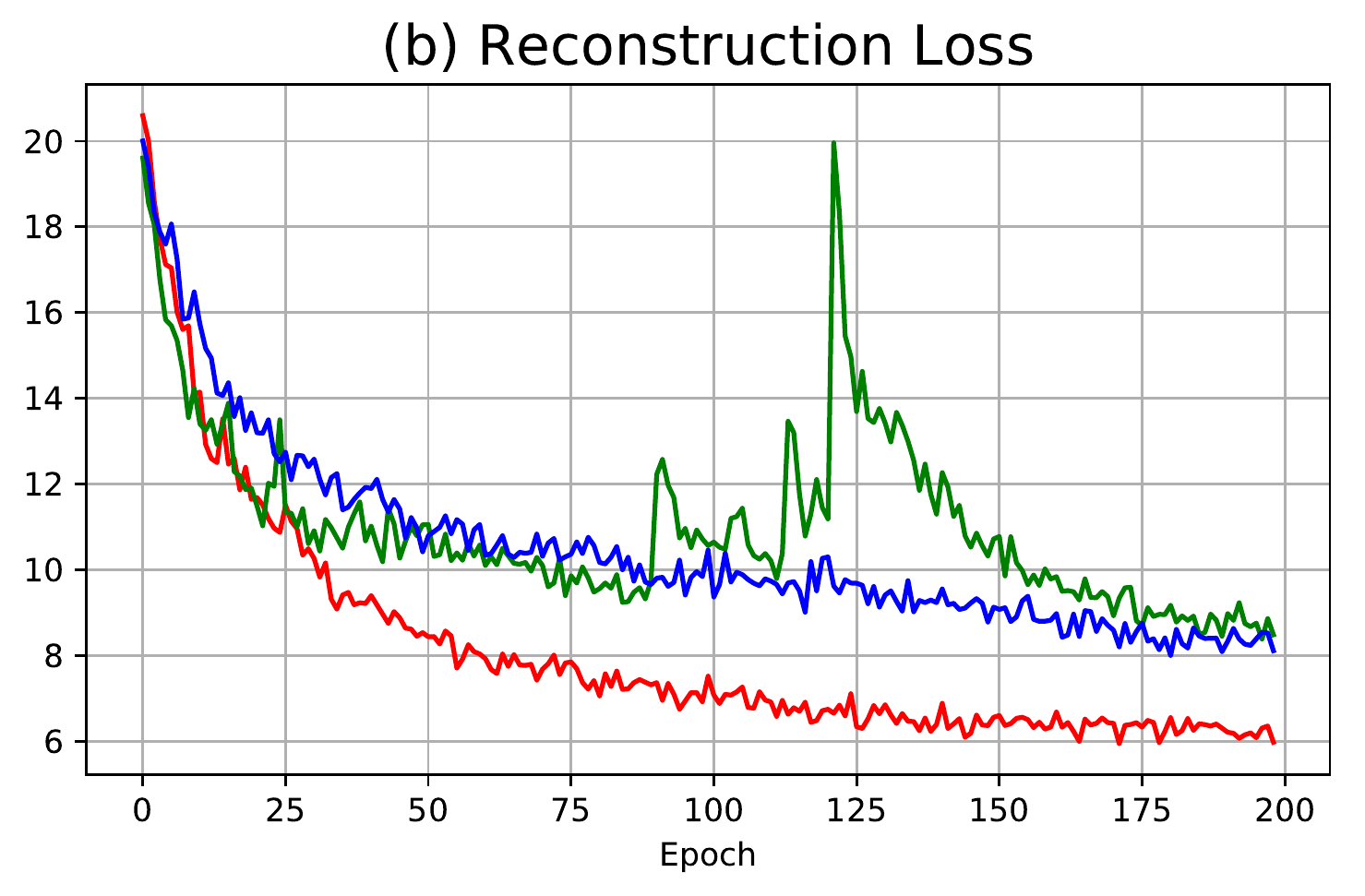}
    \end{subfigure}
    \begin{subfigure}{0.2375\textwidth}
    \centering
    \includegraphics[width=\textwidth]{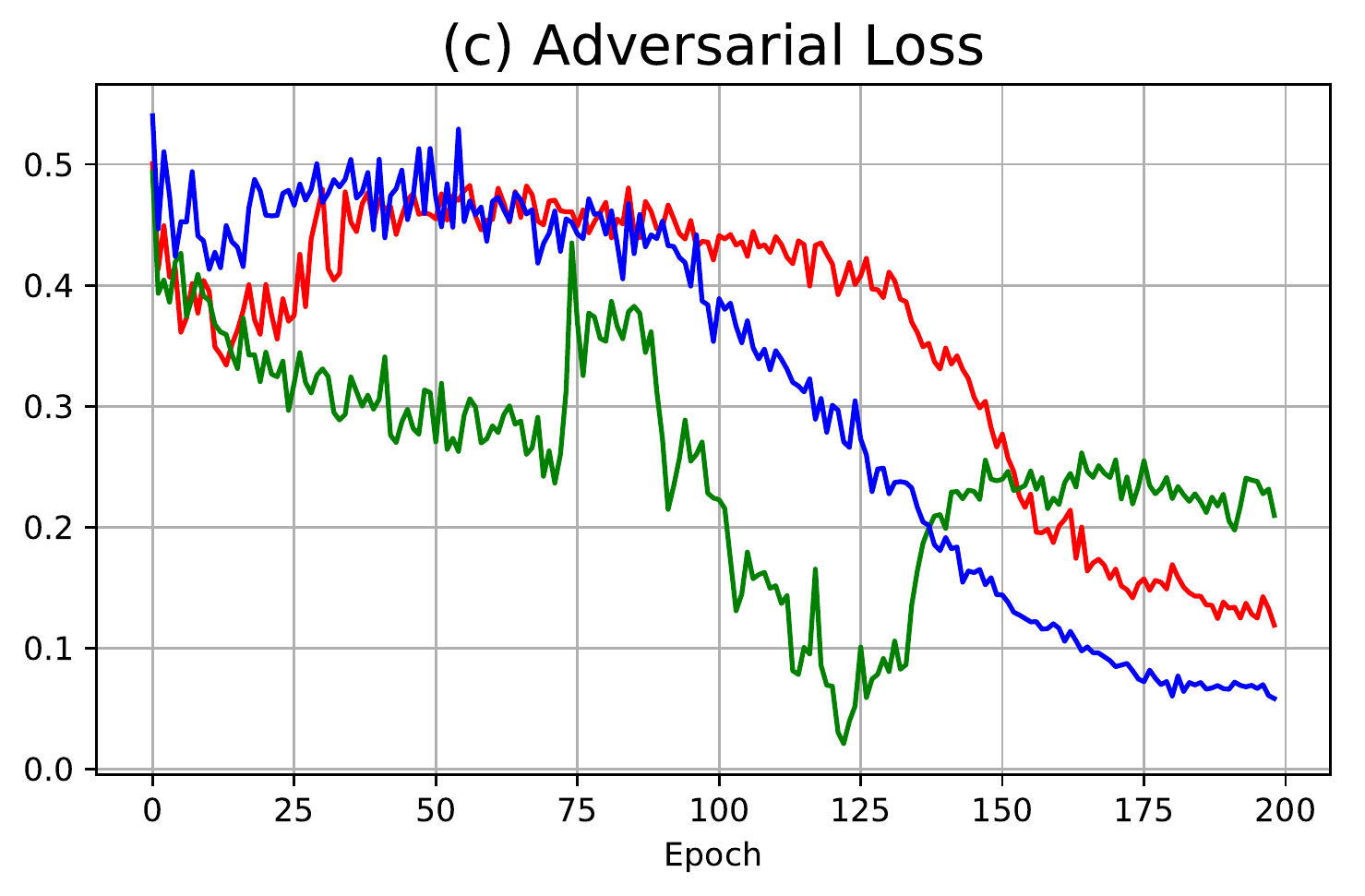}
    \end{subfigure}
    \begin{subfigure}{0.2375\textwidth}
    \centering
    \includegraphics[width=\textwidth]{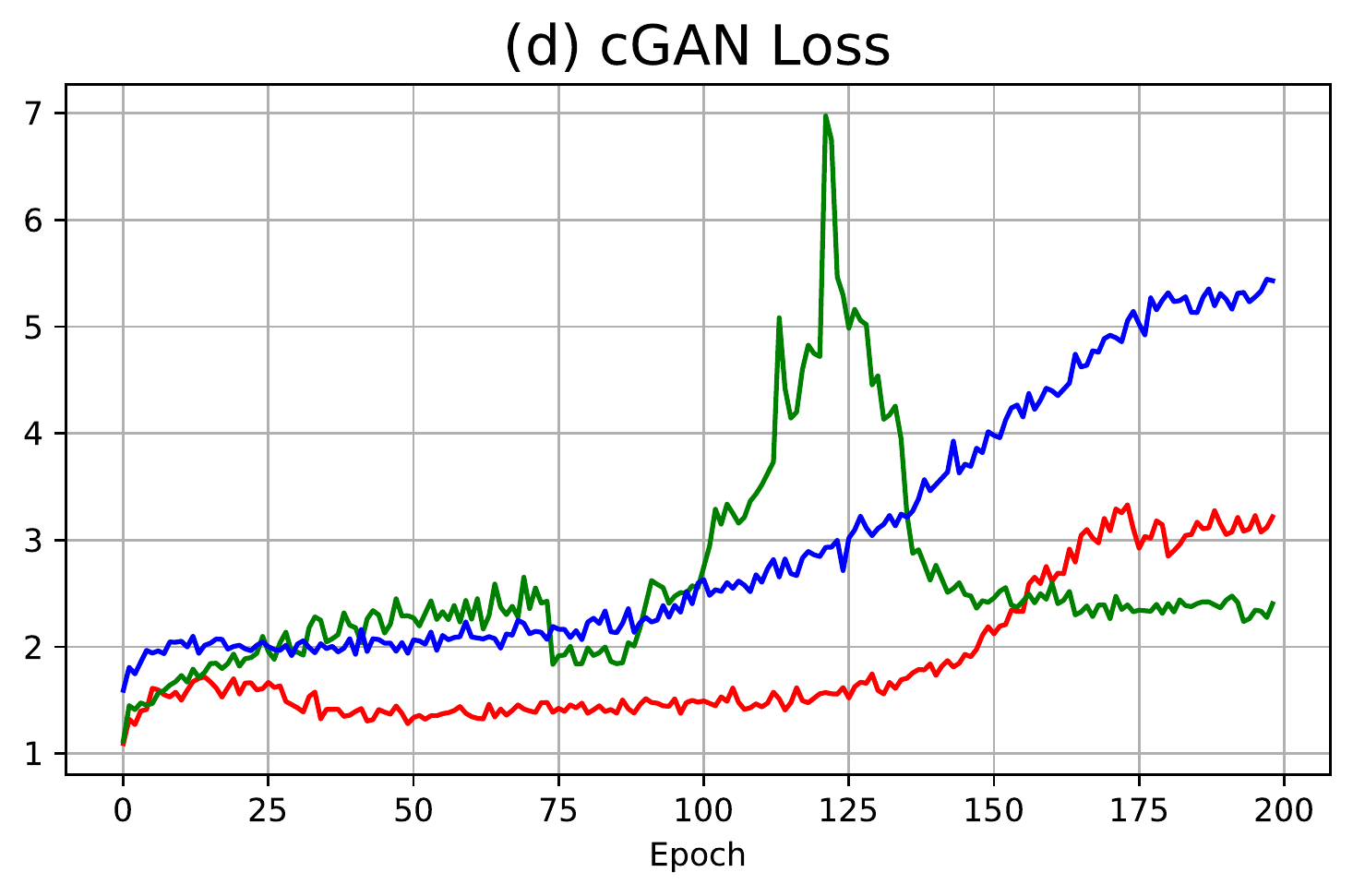}
    \end{subfigure}
    \caption{\textbf{Composition Ablation Study}. We show the values of the (a) Registration Accuracy, (b) Reconstruction loss, (c) Adversarial Loss and (d) cGAN Loss. The x-axis in all figures is the epoch number. The loss values are shown for $T\circ R$ (\textcolor{red}{red}), $R \circ T$ (\textcolor{green}{green}) and ours (\textcolor{blue}{blue}). As can be seen, the registration accuracy is best using our method. In $ T \circ R$, the reconstruction loss is the lowest, however, the registration is inaccurate because a significant portion of the registration task is implicitly performed by the translator $T$. Further, the composition $R \circ T$ is unstable because at some point, the registration network $R$ starts alternating pixels values, which is detected by the discriminator (see the dip in (c)).}
    \label{fig:geometric_preserving}
\end{figure*}

To evaluate the impact of training of $T$ and $R$ simultaneously with the two training flows proposed in Figure~\ref{fig:training_overview}, we compare the registration accuracy of our method with that of training models with either $T \circ R$ or $R \circ T$. As can be seen from Figure~\ref{fig:geometric_preserving}, training both combinations yields a substantial improvement in the registration accuracy (shown in blue), compared to each training flow (i.e., $T\circ R$ and $R\circ T$) separately. Moreover, while the reconstruction loss of $T\circ R$ (shown in read) is lowest among the three options, it does not necessarily indicate a better registration. This is because in this setting the translation network $T$ implicitly performs both the alignment and translation tasks. Conversely, when training with $R \circ T$ only (shown in green), the network $R$ is unstable and at some point it starts to alternate pixel values, essentially taking on the role of a translation network. Since $R$ is only geometry-aware by design it fails to generate good samples. This is indicated by how fast the discriminator detects that the generated samples are fake (i.e., the adversarial loss decays fast). Visual results are provided in the supplementary materials.

\begin{table}
    \centering
    \begin{center}
        \begin{tabular}{||l | c | c | c||} 
        \hline
        \backslashbox{L1 loss}{GAN loss}  & $R$ &  $T$  & Both \\ [0.5ex] 
        \hline
        $R$      & -  & $X$  & 28.15 \\
        \hline
        $T$      & 29.02 & -  & 22.03 \\
        \hline
        Both   & $X$ & $X$  & 11.01 \\
        \hline
        \end{tabular}
    \end{center}
    
        \caption{\textbf{Loss ablation results.} Columns denote modules trained with a GAN loss term. Rows denote modules trained with an L1 loss term. We do not report results in the degenerate case where only one of the modules is trained directly with any loss terms. $X$ denotes cases where the training diverges and leads to nonsensical results. For example, the result in the second row and first column represents the registration accuracy achieved when module $R$'s weights are updated with respect to the cGAN loss and module $T$ with respect to the reconstruction loss term.}
    \label{table:loss_ablation_table}
\end{table}

\textbf{Loss ablation.}
It has been shown in previous works~\cite{zhu2017toward, pix2pix2016, CycleGAN2017} that training an image-to-image translation network with both a reconstruction and an adversarial loss yields better results. In particular, the reconstruction loss stabilizes the training process and improves the vividness of the output images, while the adversarial loss encourages the generation of samples matching the real-data distribution.  

The main objective of our work is the production of a registration network. Therefore, we seek to understand the impact of both losses (reconstruction and adversarial) on the registration network. To understand the impact of each loss, we train our model with different settings: each time we fix either $R$ or $T$'s weights with respect to one of the loss functions. The registration accuracy is presented in Table \ref{table:loss_ablation_table}. Please refer to the supplementary material for qualitative results. As can be seen in these figures, training $R$ only with respect to the reconstruction loss leads to overly sharp, but unrealistic images where the deformation field creates noisy artifacts. On the other hand, training $R$ only with respect to the adversarial loss creates realistic images, but with inexact alignment. This is especially evident in Table \ref{table:loss_ablation_table} where training $R$ with respect to the reconstruction loss achieves a significant improvement in the alignment, and the best accuracy is obtained when the loss terms are both used to update all the networks weights.

\textbf{Bilateral Filtering Effectiveness}
Using bilateral filtering to weigh the smoothness loss allows us, in effect, to encourage piece-wise smoothness on the deformation map. As can be seen in Table \ref{table:bilateral_numerical}, this enhances the precision of the registration. These results suggest that using segmentation maps for controlling the smoothness loss term could be beneficial.

\begin{table}
    \centering
    \begin{center}
        \begin{tabular}{||l c c||} 
        \hline
        Method & Test Acc. &  Train Acc. \\ [0.5ex] 
        \hline\hline
        No Registration & 35.45  & 34.96 \\
        \hline
        W/O Bilateral   &  11.01 & 9.89 \\
        \hline
        With Bilateral   &  6.93 & 6.12 \\
        \hline
        \end{tabular}
    \end{center}
    \caption{\textbf{Smoothness Regularization}. Effect of bilateral filtering on registration accuracy. We show the registration accuracy on annotated test samples, and annotated train samples. As can be seen, with bilateral filtering there's less data-fitting, and the test and train accuracy relatively match.}
    \label{table:bilateral_numerical}
\end{table}

\section{Summary and Conclusions}
We presented an unsupervised multi-modal image registration technique based on image-to-image translation network. 
Our method, does not require any direct comparison between images of different modalities.
Instead, we developed a geometry preserving image-to-image translation network which allows comparing the deformed and target image using simple mono-modality metrics. The geometric preserving translation network was made possible by a novel training scheme, which alternates and combines two different flows to train the spatial transformation. We further showed that using adversarial learning, along with mono-modality metric, we are able to produce smooth and accurate registration results even when there is only little training data.

We believe that geometric preserving generators can be useful for many other applications other than image registration. In the future, we would like to continue to explore the idea of alternate training a number of layers or operators in different flows to encourage them being commutative as means to achieve certain non-trivial properties. 






\section*{Acknowledgments}
This research was supported by a generic RD program of the Israeli innovation authority, and the Phenomics consortium.

{\small
\bibliographystyle{ieee_fullname}
\bibliography{egbib}
}

\end{document}